\pdfoutput=1

\documentclass[11pt]{article}

\usepackage[]{emnlp2021}

\usepackage{times}
\usepackage{latexsym}
\usepackage{enumitem}

\usepackage[T1]{fontenc}

\usepackage[utf8]{inputenc}

\usepackage{microtype}

\usepackage{graphicx}
\usepackage{amsmath}
\usepackage{xcolor}
\usepackage{colortbl}
\usepackage{adjustbox}
\usepackage{booktabs}
\usepackage{tcolorbox}

\newcolumntype{R}[2]{%
    >{\adjustbox{angle=#1,lap=\width-(#2)}\bgroup}%
    l%
    <{\egroup}%
}
\newcommand*\rot{\multicolumn{1}{R{45}{1em}}}

\newcommand\blfootnote[1]{%
  \begingroup
  \renewcommand\thefootnote{}\footnote{#1}%
  \addtocounter{footnote}{-1}%
  \endgroup
}

\definecolor{c0}{cmyk}{0.6765,0.2017,0,0.0667} 
\newtcbox{\template}{on line,colback=c0!10,colframe=white,size=fbox,arc=3pt, box align=base,before upper=\strut,
top=-2pt, bottom=-2pt, boxrule=0pt}

\title{Exploring the Role of BERT Token Representations to\\Explain Sentence Probing Results}

\author{Hosein Mohebbi$^{\star\heartsuit}$ ~ Ali Modarressi$^{\star\heartsuit}$ ~ Mohammad Taher Pilehvar$^{\spadesuit}$ \\
  $^\heartsuit$ Iran University of Science and Technology, Iran \\
  $^\spadesuit$ Tehran Institute for Advanced Studies, Khatam University, Iran \\
  \texttt{\{hosein\_mohebbi, m\_modarressi\}@comp.iust.ac.ir}\\ 
  \texttt{mp792@cam.ac.uk}
  }

\begin{document}
\maketitle
\begin{abstract}
Several studies have been carried out on revealing linguistic features captured by BERT.
This is usually achieved by training a diagnostic classifier on the representations obtained from different layers of BERT.
The subsequent classification accuracy is then interpreted as the ability of the model in encoding the corresponding linguistic property.
Despite providing insights, these studies have left out the potential role of token representations.
In this paper, we provide a more in-depth analysis on the representation space of BERT in search for distinct and meaningful subspaces that can explain the reasons behind these probing results.
Based on a set of probing tasks and with the help of attribution methods we show that BERT tends to encode meaningful knowledge in specific token
representations (which are often ignored in standard classification setups), allowing the model to detect syntactic and semantic abnormalities, and to distinctively separate grammatical number and tense subspaces.\blfootnote{Authors marked with a star ($^\star$) contributed equally.}\footnote{Code is available at \url{https://github.com/hmohebbi/explain-probing-results}}
\end{abstract}

\section{Introduction}
Recent years have seen a surge of interest in pre-trained language models, highlighted by extensive research around BERT \cite{devlin-etal-2019-bert} and its derivatives.
One strand of research has focused on enhancing existing models with the primary objective of improving downstream performance on various NLP tasks \citep{liu2019roberta, lan2019albert, yang2019xlnet}.
Another strand analyzes the behaviour of these models with the hope of getting better insights for further developments \cite{clark-etal-2019-bert, kovaleva-etal-2019-revealing, jawahar-etal-2019-bert, tenney-etal-2019-bert, lin-etal-2019-open}.

Probing is one of the popular analysis methods, often used for investigating the encoded knowledge in language models \citep{conneau-etal-2018-cram, tenney2018you}. 
This is typically carried out by training a set of diagnostic classifiers that predict a specific linguistic property based on the representations obtained from different layers. 
Recent works in probing language models demonstrate that initial layers are responsible for encoding low-level linguistic information, such as part of speech and positional information, whereas intermediate layers are better at syntactic phenomena, such as syntactic tree depth or subject-verb agreement, while in general semantic information is spread across the entire model \citep{lin-etal-2019-open, peters2018dissecting, liu-etal-2019-linguistic, hewitt-manning-2019-structural, tenney-etal-2019-bert}. 
Despite elucidating the type of knowledge encoded in various layers, these studies do not go further to investigate the reasons behind the layer-wise behavior and the role played by token representations.
Analyzing the shortcomings of pre-trained language models requires a scrutiny beyond the mere performance (e.g., accuracy or F-score) in a given probing task.
This is particularly important as recent studies point out that the diagnostic classifier (applied to the model's outputs) might itself play a significant role in learning nuances of the task and hence suggest evaluating probes with alternative criteria \citep{hewitt-liang-2019-designing, voita-titov-2020-information, pimentel-etal-2020-information, zhu-rudzicz-2020-information}.

We extend the layer-wise analysis to the token level in search for distinct and meaningful subspaces in BERT's representation space that can explain the performance trends in various probing tasks.
To this end, we leverage the attribution method \citep{simonyan2013deep, sundararajan2017axiomatic, smilkov2017smoothgrad} which has recently proven effective for analytical studies in NLP \citep{li-etal-2016-visualizing, yuan2019interpreting, bastings-filippova-2020-elephant, atanasova-etal-2020-diagnostic, wu2021explaining, voita-etal-2021-analyzing}.
Our analysis on a set of surface, syntax, and semantic probing tasks \citep{conneau-etal-2018-cram} shows that BERT usually encodes the knowledge required for addressing these tasks within specific token representations, particularly at higher layers.
For instance, we found that sentence-ending tokens (e.g., ``\textsc{[sep]}'' and ``.'') are mostly responsible for carrying positional information through layers, or when the input sequence undergoes a re-ordering the alteration is captured by specific token representations, e.g., by the swapped tokens or the coordinator between swapped clauses.
Also, we observed that the \texttt{\#\#s} token is mainly responsible for encoding noun number and verb tense information, and that BERT clearly distinguishes the two usages of the token in higher layer representations.

\section{Related Work}
\label{sec:relatedWork}

\paragraph{Probing.}
Several analytical studies have been conducted to examine the capacities and weaknesses of BERT, often by means of probing layer-wise representations \citep{lin-etal-2019-open, goldberg2019assessing, liu-etal-2019-linguistic, jawahar-etal-2019-bert, tenney-etal-2019-bert}.
Particularly, \citet{jawahar-etal-2019-bert} leveraged the probing framework of \citet{conneau-etal-2018-cram} to show that BERT carries a hierarchy of linguistic information, with surface, syntactic, and semantic features respectively occupying initial, middle and higher layers.
In a similar study, \citet{tenney-etal-2019-bert} employed the edge probing tasks defined by \citet{tenney2018you} to show the hierarchy of encoded knowledge through layers.
Moreover, they observed that while most of the syntactic information can be localized in a few layers, semantic knowledge tends to spread across the entire network. 
Both studies were aimed at discovering the extent of linguistic information encoded across different layers.
In contrast, in this paper we explore the role of token representations in the final performance.
More recently, \citet{klafka-ettinger-2020-spying} investigated the extent of information that can be recovered from each word representation in a sentence about the other words. 
Apart from using different probing tasks and methodologies, most notably they relied solely on classifier's performance score, whereas we make conclusion based on the most contributed token representations.

\paragraph{Representation subspaces.}
In addition to layer-wise representations, subspaces that encode specific linguistic knowledge, such as syntax, have been a popular area of study.
By designing a structural probe, \citet{hewitt-manning-2019-structural} showed that there exists a linear subspace that approximately encodes all syntactic tree distances. 
In a follow-up study, \citet{chi-etal-2020-finding} showed that similar syntactic subspaces exist for languages other than English in the multilingual BERT and that these subspaces are shared among languages to some extent. 
This corroborated the finding of \citet{pires-etal-2019-multilingual} that multilingual BERT has common subspaces across different languages that capture various linguistic knowledge.

As for semantic subspaces, \citet{wiedemann-etal-2019-does} showed that BERT places the contextualized representations of polysemous words into different regions of the embedding space, thereby capturing sense distinctions. 
Similarly, \citet{reif2019visualizing}
studied BERT's ability to distinguish different word senses in different contexts. 
Using the probing approach of \citet{hewitt-manning-2019-structural}, they also found that there exists a linear transformation under which distances between word embeddings correspond to their sense-level relationships.
Our work extends these studies by revealing other types of surface, syntactic, and high-level semantic subspaces and linguistic features using a pattern-finding approach on different types of probing tasks.

\paragraph{Attribution methods.}
Recently, there has been a surge of interest in using attribution methods to open up the blackbox and explain the decision makings of pre-trained language models, from developing methods and libraries to visualize inputs' contributions \citep{ribeiro-etal-2016-trust, han-etal-2020-explaining, wallace-etal-2019-allennlp, tenney-etal-2020-language} to applying them into fine-tuned models on downstream tasks \citep{atanasova-etal-2020-diagnostic, wu2021explaining, voita-etal-2021-analyzing}. In particular, \citet{voita-etal-2021-analyzing} adopted a variant of Layer-wise Relevance Propagation \cite{Bach-etal-2015} to evaluate the relative contributions of source and target tokens to the generation process in Neural Machine Translation predictions. To our knowledge, this is the first time that attribution methods are employed for layer-wise probing of pre-trained language models.

\section{Methodology}
\label{sec:methodology}

Our analytical study was mainly carried out on a set of sentence-level probing tasks from SentEval \citep{conneau-kiela-2018-senteval}.
The benchmark consists of several single-sentence evaluation tasks. Each task provides 100k instances for training and 10k for test, all balanced across target classes. 
We used the test set examples for our evaluation and in-depth analysis.
Following the standard procedure for this benchmark, we trained a diagnostic classifier for each task.
The classifier takes sentence representations as its input and predicts the specific property intended for the corresponding task.

In what follows in this section, we first describe how sentence representations were computed in our experiments.
Then, we discuss our approach for measuring the attribution of individual token representations to classifier's decision.

\subsection{Sentence Representation}
\label{sec:sentence-rep}
For computing sentence representations for layer $l$, we opted for a simple unweighted averaging ($h_{Avg}^l$) of all input tokens (except for padding and \textsc{[cls]} token).
This choice was due to our observation that the mean pooling strategy retains or improves \textsc{[cls]} performance in most layers in our probing tasks (cf. Appendix \ref{sec:appMeanVsCLS} for more details). 
This corroborates the findings of \citet{reimers-gurevych-2019-sentence} who observed a similar trend on sentence similarity and inference tasks.
Moreover, the mean pooling strategy simplifies our measuring of each token's attribution, discussed next.

Our evaluations are based on the pre-trained BERT (base-uncased, 12-layer, 768-hidden size, 12-attention head, 110M parameters) obtained from the HuggingFace's Transformers library \citep{wolf-etal-2020-transformers}.
We followed the recommended hyperparameters by \citet{jawahar-etal-2019-bert} to train the diagnostic classifiers for each layer.
In addition to BERT, we carried out our evaluations on RoBERTa  \citep[base, 125M parameters]{liu2019roberta}. However, we observed highly similar patterns for the two models. Hence, we only report results for the BERT model.

\subsection{Gradient-based Attribution Method}
\label{sec:attrMethod}

We leveraged a gradient-based attribution method in order to enable an in-depth analysis of layer-wise representations with the objective of explaining probing performances.
Specifically, we are interested in computing the attribution of each input token to the output labels.
This is usually referred to as the \textit{saliency} score of an input token to classifier's decision.
Note that using attention weights for this purpose can be misleading given that raw attention weights do not necessarily correspond to the importance of individual token representations \citep{serrano-smith-2019-attention, jain-wallace-2019-attention, abnar-zuidema-2020-quantifying, kobayashi-etal-2020-attention}.

Using gradients for attribution methods has been a popular option in neural networks, especially for vision \citep{simonyan2013deep, sundararajan2017axiomatic, smilkov2017smoothgrad}. 
Images are constructed from pixels; hence, computing their individual attributions to a given class can be interpreted as the spatial support for that class \citep{simonyan2013deep}. 
However, in the context of text processing, input tokens are usually represented by vectors; hence, raw feature values do not necessarily carry any specific information. 
\citet{li-etal-2016-visualizing}'s solution to this problem relies on the gradients over the inputs. Let $w_c$ be the derivative of class $c$'s output logit ($y_c$) with respect to the $k$-th dimension of the input embedding ($h[k]$): 
\begin{equation}
\label{eq:sensitivity}
    w_c(h[k]) = \frac{\partial y_c}{\partial h[k]}
\end{equation}
This gradient can be interpreted as the sensitivity of class $c$ to small changes in $h[k]$. 
To have this at the level of words (or tokens), \citet{li-etal-2016-visualizing} suggests using the average of the absolute values of $w_c(h[k])$ over all of the $d$ dimensions of the embedding:
\begin{equation}
\label{eq:avg_sensitivity}
    Score_c(h) = \frac{1}{d}\sum_{k=1}^{d}|w_c(h[k])|
\end{equation}
Although the absolute value of gradients could be employed for understanding and visualizing the contributions of individual words, these values can only express the sensitivity of the class score to small changes without information about the direction of contribution \citep{yuan2019interpreting}.
We adopt the method of \citet{yuan2019interpreting} for our setting and compute the saliency score for the $i^{\text{th}}$ representation in layer $l$, i.e.,  $h^l_i$, as:
\begin{equation}
\label{eq:score}
    Score_c(h^l_i) = \frac{\partial y^l_c}{\partial h_{Avg}^l} \cdot h^l_i
\end{equation}
where $y^l_c$ denotes the probability that the classifier assigns to class $c$ based on the $l^{\text{th}}$-layer representations.
Given that our aim is to explain the representations (rather than evaluating the classifier), we set $c$ in Equation \ref{eq:score} as the correct label.
This way, the scores reflect the contributions of individual input tokens in a sentence to the classification decision.


In what follows in the paper, we use the analysis method discussed in this section to find those tokens that play the central role in different surface (Section \ref{sec:sentleng}), syntactic (Sections \ref{sec:tense-task} and \ref{bshift_experiment}) and semantic (Section \ref{sec:phrase-inv}) probing tasks. 
Based on these tokens we then investigate the reasons behind performance variations across layers.

\section{Sentence Length}
\label{sec:sentleng}
In this surface-level task we probe the representation of a given sentence in order to estimate its size, i.e., the number of words (not tokens) in it. 
To this end, we used SentEval's \textbf{SentLen} dataset, but changed the formulation from the original classification objective to a regression one which allows a better generalization due to its fine-grained setting. 
The diagnostic classifier receives average-pooled representation of a sentence (cf. Section \ref{sec:sentence-rep}) as input and outputs a continuous number as an estimate for the input length.

Given that the ability to encode the exact length of input sentences is not necessarily a critical feature, we do not focus on layer-wise performance and instead discuss the reason behind the performance variations across layers.
To this end, we calculated the absolute saliency scores for each input token in order to find those tokens that played pivotal role while estimating sentence length.

Rounding the regressed estimates and comparing them with the gold labels in the test set, we can observe a significant performance drop from 0.91 accuracy in the first layer to 0.44 in the last layer (cf. Appendix \ref{sec:app_avg_rep} for details). 
This decay is not surprising given that the positional encodings, which are added to the input embeddings in BERT and are deemed to be the main players for such a position-based task, get faded through layers \citep{voita-etal-2019-bottom}.

\begin{figure}[t!]
    \includegraphics[width=\linewidth]{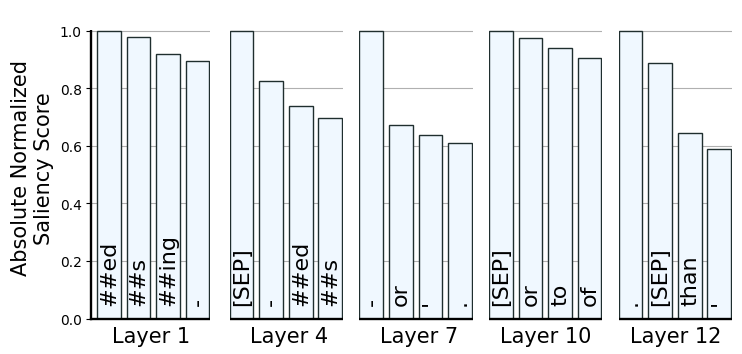}
    \caption{Absolute normalized saliency scores for the top-4 most attributed (high frequency, $>128$) tokens across five different layers.\footnotemark}
    \label{fig:senlen_bar}
\end{figure}

\footnotetext{Full figures (for all layers) are available in Appendix
\ref{sec:FullApp}}

\begin{figure}[t!]
    \includegraphics[width=\linewidth]{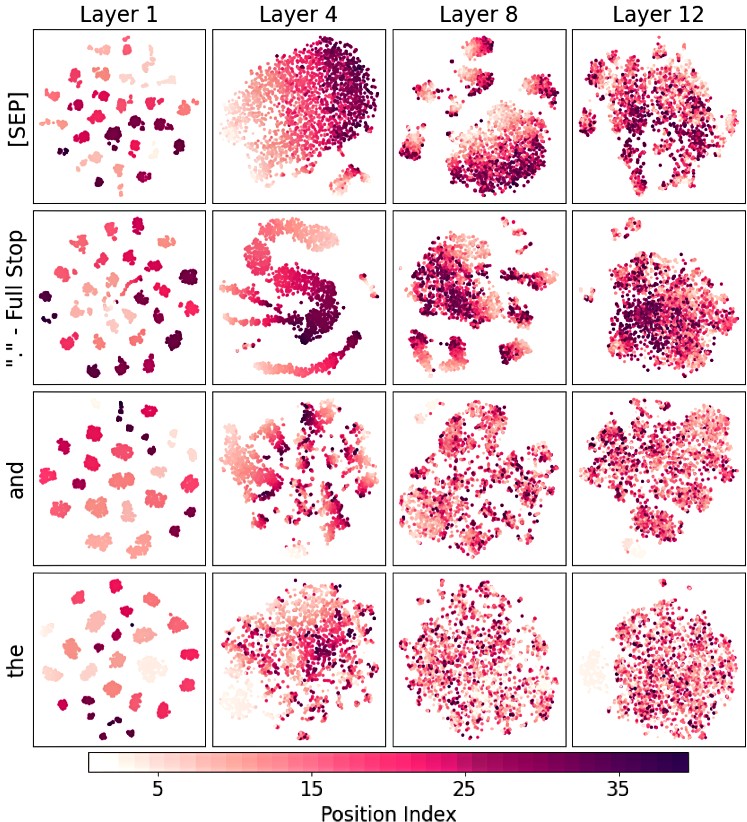}
    \caption{t-SNE plots of the representations of four selected high frequency tokens (``\textsc{[sep]}'', ``.'' full stop, ``the'', ``and'') in different sentences. Colors indicate the corresponding token's position in the sentence (darker colors means higher position index). Finalizing tokens (e.g., ``\textsc{[sep]}'', ``.'') preserve distinct patterns in final layers, indicating their role in encoding positional information, while other (high frequency) tokens exhibit no such behavior.}
    \label{fig:tokenpositions}
\end{figure}

\paragraph{Sentence ending tokens retain positional information.}
Figure \ref{fig:senlen_bar} shows tokens that most contributed to the probing results across different layers according to the attribution analysis.
Finalizing tokens (e.g. ``\textsc{[sep]}'' and ``.'') are the main contributors in the higher layers.
We further illustrate this in Figure \ref{fig:tokenpositions} in which we compare the representations of a finalizing token with those of another frequent non-finalizing token.
Clearly, positioning information is lost throughout layers in BERT; however, finalizing tokens partially retain this information, as visible from distinct pattern in higher layers.



\section{Verb Tense and Noun Number}
\label{sec:tense-task}

This analysis inspects BERT representations for grammatical number and tense information. 
For this experiment we used the \textbf{Tense} and \textbf{ObjNum} tasks\footnotemark\footnotetext{We will not discuss the SubjNum results, since we observed significant labeling issues (See Appendix \ref{sec:subjNumMislabel}) that could affect our conclusions. This can also explain low human performance reported by \citet{conneau-etal-2018-cram} on this task.}: the former checks whether the main-clause verb is labeled as present or past\footnotemark\footnotetext{In Tense task, each sentence may include multiple verbs, subjects, and objects, while the label is based on the main clause \citep{conneau-etal-2018-cram}.}, whereas the latter classifies the object according to its number, i.e., singular or plural \citep{conneau-etal-2018-cram}.
On both tasks, BERT preserves a consistently high performance ($>$ 0.82 accuracy) across all layers (cf. Appendix \ref{sec:app_avg_rep} for more details).

\begin{figure*}[t!]
    \centering
        \centering
        \includegraphics[width=\linewidth]{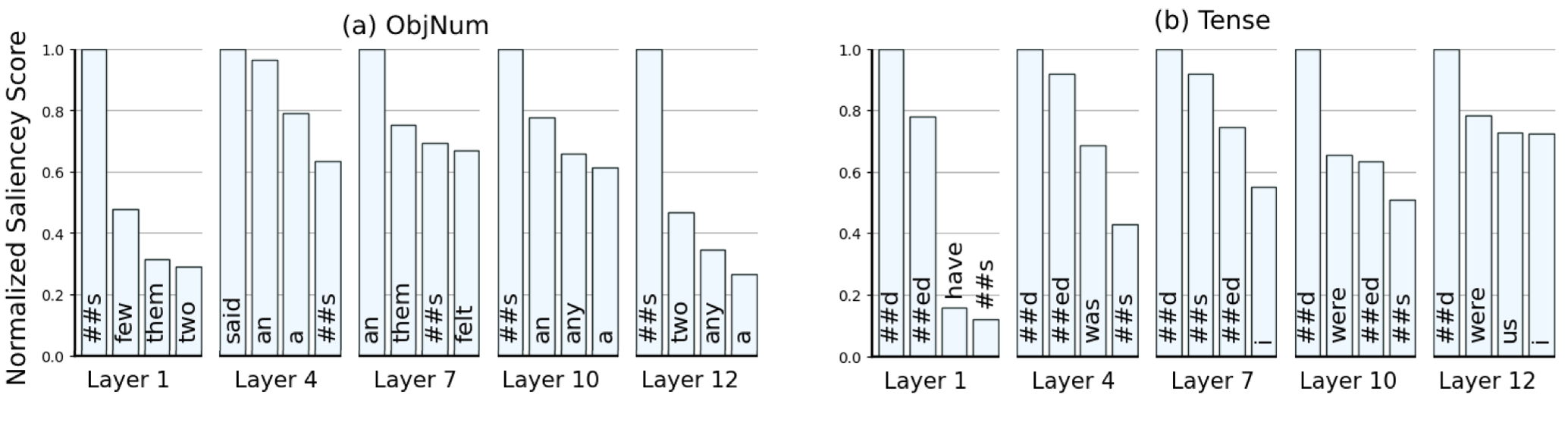}
        \caption{The top-4 most attributed (high freq.) tokens across five different layers for the ObjNum and Tense tasks.}
        \label{fig:tenseObjNum}
\end{figure*}

\begin{figure*}[t!]
    \includegraphics[width=\linewidth]{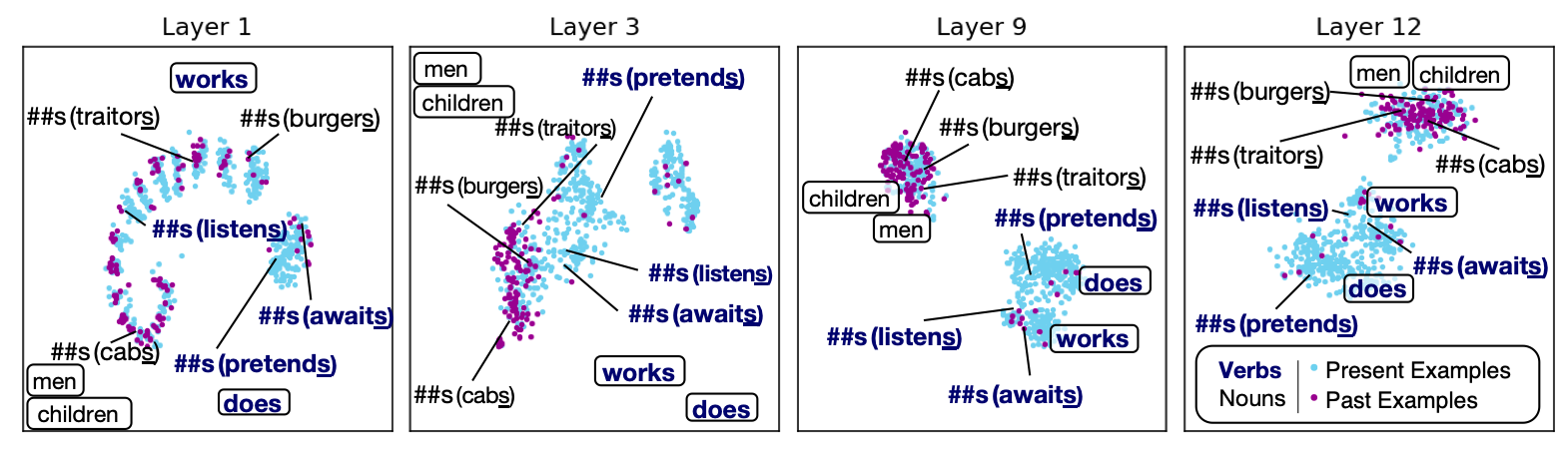}
    \caption{t-SNE plots of the layer-wise representations of the \texttt{\#\#s} token in different sentences. Colors indicate whether the token occurred in present- or past-labeled sentence in the Tense task (see Section \ref{sec:tense-task}). For the sake of comparison, we also include two present verbs without the \texttt{\#\#s} token\footnotemark (i.e., \textit{does} and {\it works}) and two irregular plural nouns (i.e., \textit{men} and \textit{children}), in rounded boxes. The distinction between the two different usages of the token (noun number as well as the tense information) is clearly encoded in higher layer contextualized representations. As plural nouns can appear in both past- and present-labeled examples, the cluster belongs to the plural form of \texttt{\#\#s} token in higher layers may contain both types of examples.} 
    \label{fig:sPresentPast}
\end{figure*}

\paragraph{Articles and ending tokens (e.g., \texttt{\#\#s} and \texttt{\#\#ed}) are key playmakers.}
Attribution analysis, illustrated in Figure \ref{fig:tenseObjNum}(a), reveals that article words (e.g., ``a'' and ``an'') and the ending \texttt{\#\#s} token, which makes out-of-vocab plural words (or third person present verbs), are among the most attributed tokens in the ObjNum task.
This shows that these tokens are mainly responsible for encoding object's number information across layers. 
As for the Tense task, Figure \ref{fig:tenseObjNum}(b) shows a consistently high influence from verb ending tokens (e.g., \texttt{\#\#ed} and \texttt{\#\#s}) across layers which is in line with performance trends for this task and highlights the role of these tokens in preserving verb tense information.

\footnotetext{Tokens that were not split by the tokenizer.}
\paragraph{\texttt{\#\#s} --- Plural or Present?}
The \texttt{\#\#s} token proved influential in both tense and number tasks.
The token can make a verb into its simple present tense (e.g., read $\rightarrow$ reads) or transform a singular noun into its plural form (e.g., book $\rightarrow$ books). 
We further investigated the representation space to check if BERT can distinguish this nuance.
Results are shown in Figure \ref{fig:sPresentPast}: after the initial layers, BERT recognizes and separates these two forms into two distinct clusters (while BERT's tokenizer made no distinction among different usages). 
Interestingly, we also observed that other present/plural tokens that did not have the \texttt{\#\#s} token aligned well with these subspaces.

\section{Inversion Abnormalities}
For this set of experiments, we opted for SentEval's Bi-gram Shift and Coordination Inversion tasks which respectively probe model's ability in detecting syntactic and semantic abnormalities.
The goal of this analysis was to to investigate if BERT encodes inversion abnormality in a given sentence into specific token representations.

\subsection{Word-level inversion}
\label{bshift_experiment}
Bi-gram Shift (\textbf{BShift}) checks the ability of a model to identify whether two adjacent words within a given sentence have been inverted \citep{conneau-etal-2018-cram}. 
Probing results shows that the higher half layers of BERT can properly distinguish this peculiarity (Figure \ref{fig:bshift_sim}). Similarly to the previous experiments, we leveraged the gradient attribution method to figure out those tokens that were most effective in detecting the inverted sentences.
Given that the dataset does not specify the inverted words, we reconstructed the inverted examples by randomly swapping two consecutive words in the original sentences of the test set, excluding the beginning of the sentences and punctuation marks as stated in \citep{conneau-etal-2018-cram}.

\begin{figure}[!t]
    \includegraphics[width=\linewidth]{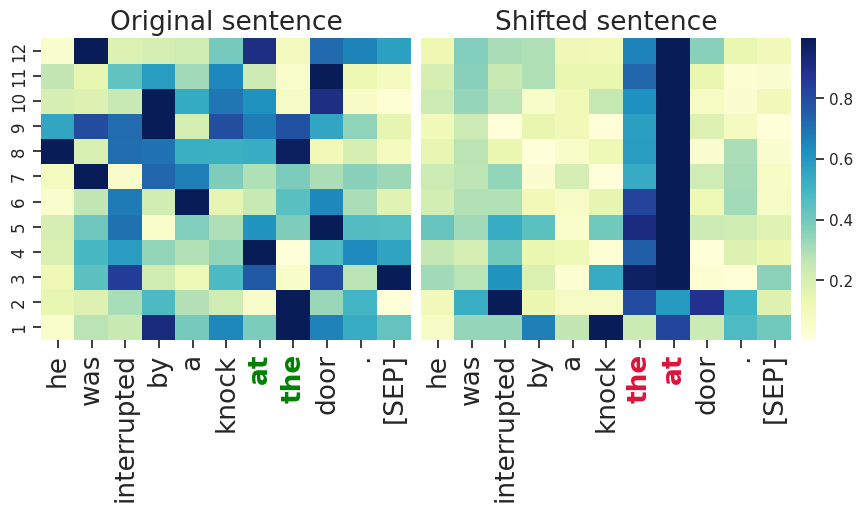}
    \caption{Normalized layer-wise attribution scores for a randomly sampled sentence from the test set (left). The right figure shows how the attribution scores changed when two words (``at'' and ``the'') from the original sentence were inverted.}
    \label{fig:bshift_heatmap}
\end{figure}

\subsection{Results}
Our attribution analysis shows that swapping two consecutive words in a sentence results in a significant boost in the attribution scores of the inverted tokens. 
As an example, Figure \ref{fig:bshift_heatmap} depicts attribution scores of each token in a randomly sampled sentence from the test set across different layers. 
The classifier distinctively focuses on the token representations for the shifted words (Figure \ref{fig:bshift_heatmap} right), while no such patterns exists for the original sentence (Figure \ref{fig:bshift_heatmap} left).

\begin{figure}[t!]
    \includegraphics[width=7cm,height=7cm,keepaspectratio]{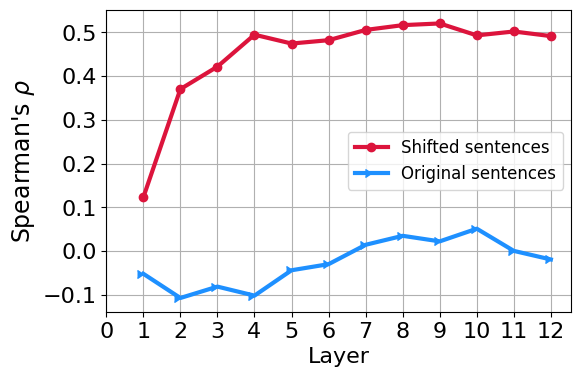}
    \caption{Spearman's $\rho$ correlation of gradient attribution scores with the mask array $M$ (a one-hot indicating shifted indices), averaged on all examples across all layers. 
    High correlations indicate model's increased sensitivity to the shifted tokens, a trend which is not seen in the original sentences.}
    \label{fig:bshift_corr}
\end{figure}

\label{sec:patternAnalysis}
To verify if this observation holds true for other instances in the test set, we carried out the following experiment.
For each given sequence $X$ of $n$ tokens, we defined a boolean mask $M = [m_1, m_2, ... m_n]$ which denotes the position of the inversion according to the following condition:
\begin{equation}
m_i = 
  \begin{cases}
    1, & x_i \in  $~~V$ \\
    0, & \textrm{otherwise} \
  \end{cases}
\end{equation}
\noindent where $V$ is the set of all tokens in the shifted bi-gram ($|V|\ge2$, given BERT's sub-word tokenization).
Then we computed the Spearman's rank correlation coefficient of the attribution scores with $M$ for all examples in the test set.
Figure \ref{fig:bshift_corr} reports mean layer-wise correlation scores. 
We observe that in altered sentences the correlation significantly grows over the first few layers which indicates model's increased sensitivity to the shifted tokens.

\begin{figure}[t!]
    \includegraphics[width=\linewidth]{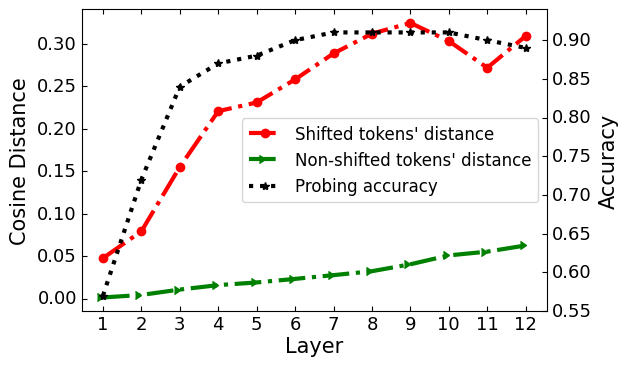}
    \caption{Average cosine distances of shifted tokens (and other tokens) to themeselves, before and after inversion in the BShift task. The trend for the shifted token distances highly correlates with that of probing performance, supporting our hypothesis of BERT encoding abnormalities in the shifted tokens.}
    \label{fig:bshift_sim}
\end{figure}

\begin{figure*}[pt]
    \includegraphics[width=\linewidth]{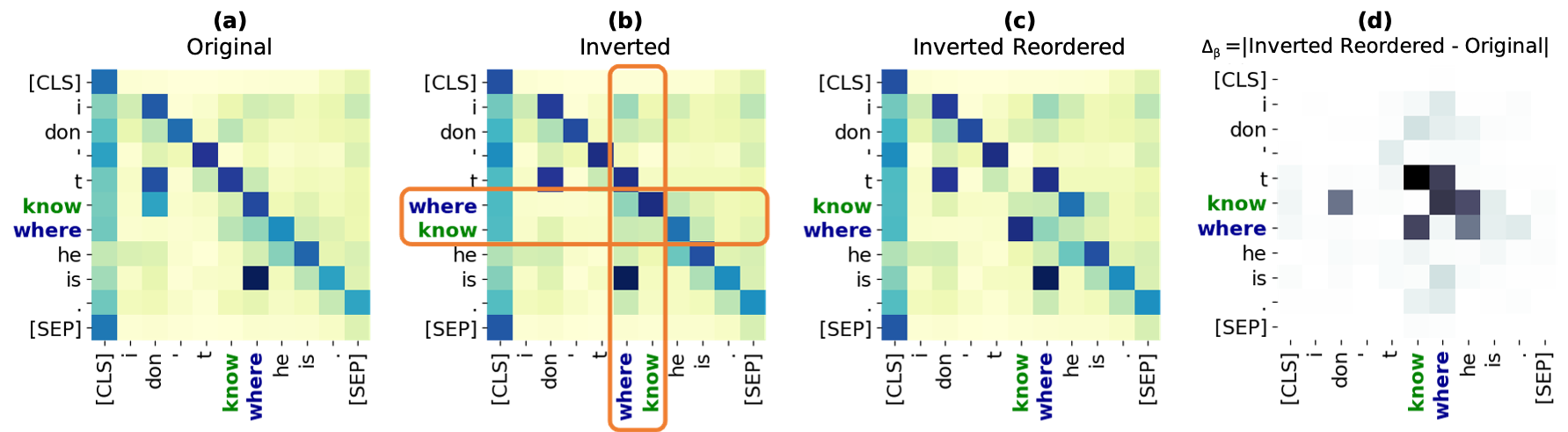}
    \caption{Evaluating the $\beta$ map\footnotemark for a single example in a specific layer (layer = 3). After computing the map for the original (a) and inverted (b) forms of the sentence, to compute the $\Delta_{\beta}$ map we need to reorder the inverted map. The corresponding columns and rows for the inverted words (orange boxes) are swapped to re-construct the original order (c). The $\Delta_{\beta}$ map (d) is the magnitude of the point-wise difference between the re-ordered and the original maps. The $\Delta_{\beta}$ map for this example clearly shows that most of the changes have occurred within the bi-gram inversion area. All values are min-max normalized.}
    \label{fig:bshift_att_explain}
\end{figure*}

We hypothesize that BERT implicitly encodes abnormalities in the representation of shifted tokens. 
To investigate this, we computed the cosine distance of each token to itself in the original and shifted sentences.
Figure \ref{fig:bshift_sim} shows layer-wise statistics for both shifted and non-shifted tokens.
Distances between the shifted token representations aligns well with the performance trend for this probing task (also shown in the figure).

\subsubsection{Attention-norm behavior on bi-gram inversion}
Our observation implies that BERT somehow encodes oddities in word order in the representations of the involved tokens. 
To investigate the root cause of this, we took a step further and analyzed the building blocks of these representations, i.e., the self-attention mechanism.
To this end, we made use of the norm-based analysis method of \citet{kobayashi-etal-2020-attention} which incorporates both attention weights and transformed input vectors (the value vectors in the self-attention layer). 
The latter component enables a better interpretation at the token level.
This norm-based metric $||\sum\alpha f(x)||$--for the sake of convenience we call it \textbf{attention-norm}--is computed as the vector-norm of the $i^{\text{th}}$ token to the $j^{\text{th}}$ token over all attention heads ($H = 12$) in each layer $l$:
\begin{equation}
\label{eq:vector-norm}
    \beta_{i,j}^l = ||\sum_{head=1}^{\text{H}}\alpha_{i,j}^{head, l} f^{head, l}(\mathbf{h}^l_j)||
\end{equation}
where $\alpha_{i,j}$ is the attention weight between the two tokens and $f^{head, l}(x)$ is a combination of the value transformation in layer $l$ of the head and the matrix which combines all heads together (see \citet{kobayashi-etal-2020-attention}'s paper for more details).

\begin{figure}[t!]
    \includegraphics[width=\linewidth]{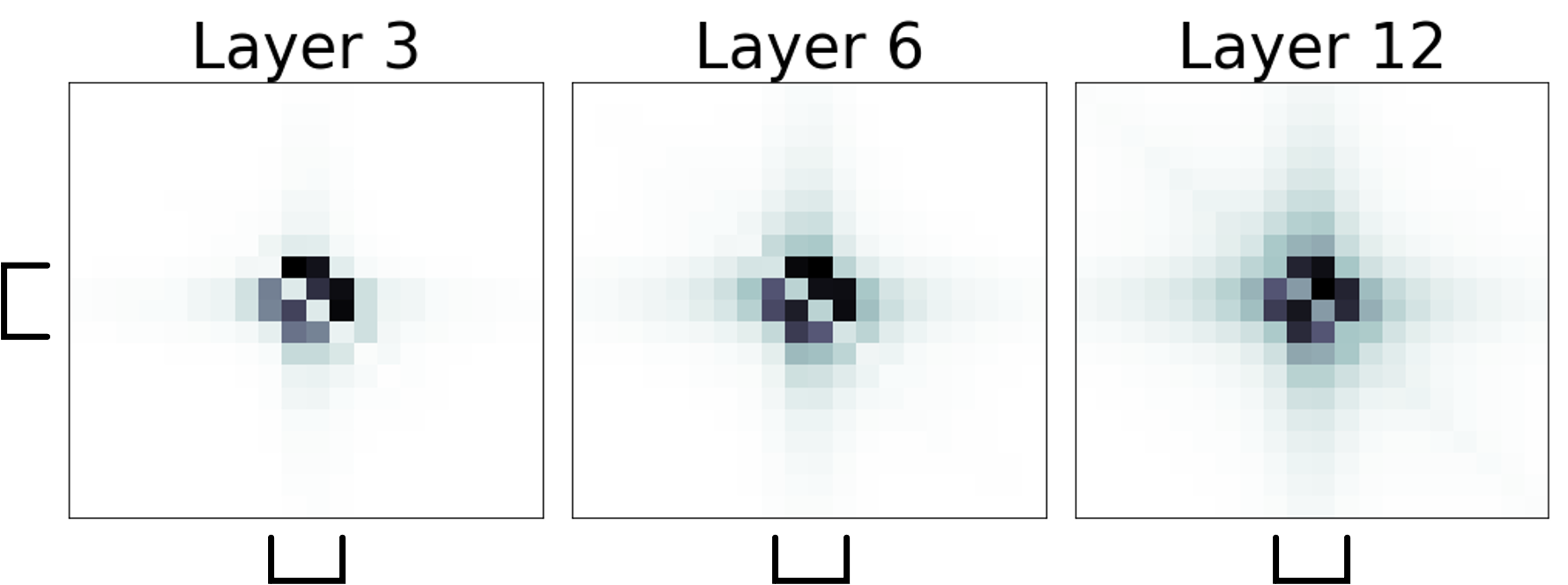}
    \caption{A cumulative view of the attention-norm changes ($\Delta_{\beta^l}$) centered around the bi-gram position (the approximate bi-gram position is marked on each figure). Each plot indicates the cumulative layer-wise changes until a specific layer.
    Each row indicates the corresponding token's attention-norms to every token in the sentence (including itself). Although the changes slightly spread out to the other tokens as we move up to higher layers, they mostly occur in the bi-gram area. Given BERT's contextualization mechanism, variations in attention-norms in each row directly result in a change in the corresponding token's representation. Therefore, the tokens in the bi-gram undergo most changes in their representations.}
    \label{fig:BShiftAttention}
\end{figure}

We computed the attention-norm map in all layers, for both the original and shifted sentence.\footnotetext{The value of the cell $\beta_{ij}$ ($i^{th}$ row, $j^{th}$ column) in the map denotes the attention-norm of the $i^{th}$ token to the $j^{th}$ token. The contextualized embedding for the $i^{th}$ token is constructed based on a weighted combination of their corresponding attention-norms in the $i^{th}$ row.}
To be able to compare these two maps, we re-ordered the shifted sentence norms to match the original order. 
The magnitude of the difference between the original and the re-ordered map $\Delta_{\beta^l}$ shows the amount of change in each token's attention-norm to each token.
Figure \ref{fig:bshift_att_explain} illustrates this procedure for a sample instance. 
Given that bi-gram locations are different across each instance, to compute an overall $\Delta_{\beta^l}$ we centered each map based on the position of the inversion. 
As a result of this procedure, we obtained a $\Delta_{\beta^l}$ map for each layer and for all examples.
Centering and averaging all these maps across layers produced Figure \ref{fig:BShiftAttention}.

Figure \ref{fig:BShiftAttention} indicates that after inverting a bi-gram, both words' attention-norms to their neighboring tokens change and this mostly affects their own representations rather than others. This observation suggests that the distinction formed between the representations of the original and shifted tokens, as was seen in Figure \ref{fig:bshift_sim}, can be rooted back to the changes in attention heads' patterns.

\subsection{Phrasal-level inversion}
\label{sec:phrase-inv}

The Coordination Inversion (\textbf{CoordInv}) task is a binary classification that contains sentences with two coordinated clausal conjoints (and only one coordinating conjunction). 
In half of the sentences the clauses' order is inverted and the goal is to detect malformed sentences at phrasal level
\citep{conneau-etal-2018-cram}. 
Since the phrasal-level inversion does not alter the syntax structure of the sentence, the task could be considered as a semantic one \cite{conneau-etal-2018-cram}. For an example:
\begin{itemize}[label={}]
    \item \template{\textsf{\small the glass broke and i cut myself . $\rightarrow$ Original}}
    \item \template{\textsf{\small i cut myself and the glass broke . $\rightarrow$ Inverted}}
\end{itemize}
\noindent While both sentences are syntactically correct, we should rely on the meaning of the sequence of the events in order to detect the abnormality in the second sentence.

BERT's performance on this task increases through layers and then slightly decreases in the last three layers.
We observed that the attribution scores for ``but'' and ``and'' coordinators to be among the highest (see Appendix \ref{sec:FullApp}) and that these scores notably increase through layers.
We hypothesize that BERT might implicitly encodes phrasal level abnormalities in specific token representations.

\begin{figure}[!t]
    \includegraphics[width=\linewidth]{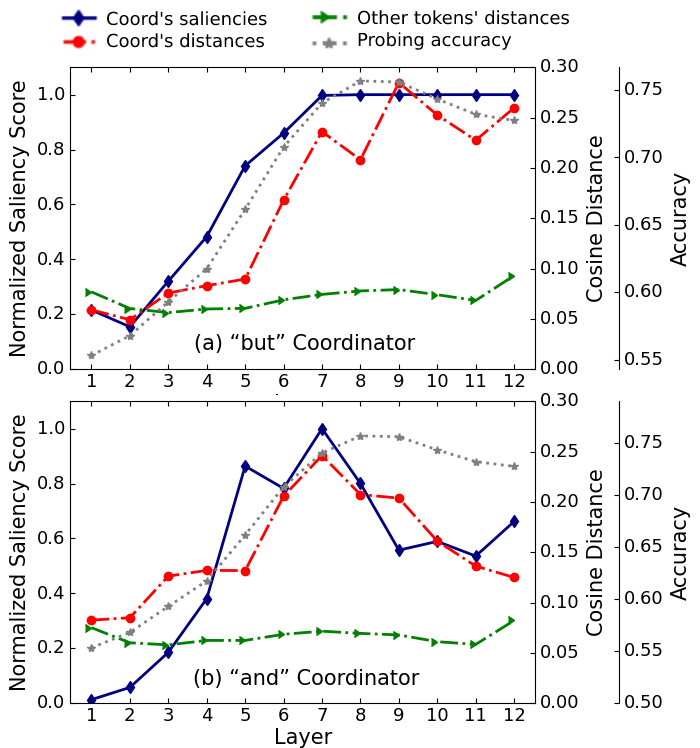}
    \caption{Averaged cosine distances between coordinators in the original and inverted sentences. We also show the normalized saliency scores for the coordinators across layers  which correlate with the performance scores of the task. The distance curve for other tokens is a baseline to highlight that the representation of coordinators significantly change after inversion.}
    \label{fig:but_and_sim_saliency}
\end{figure}

\paragraph{Odd Coordinator Representation.}
To verify our hypothesis, we filtered the test set to ensure all sentences contain either a ``but'' or an ``and'' coordinator\footnote{9,883 of the 10K examples in the test set meet this condition.}.
We reconstructed the original examples by inverting the order of the two clauses in the inverted instances since no sentence appears with both labels in the dataset.
Feeding this to BERT, we extracted token representations and computed the cosine distance between the representations of each token in the original and inverted sentences.
Figure \ref{fig:but_and_sim_saliency} shows these distances, as well as the normalized saliency score for coordinators (averaged on all examples in each layer), and layer-wise performance for the CoordInv probing task.
Surprisingly, all these curves exhibit a similar trend.
As we can see, when the order of the clauses are inverted, the representations of the coordinators ``but'' or ``and'' play a pivotal role in making sentence representations distinct from one another while there is nearly no change in the representation of other words.
This observation implies that BERT somehow encodes oddity in the coordinator representations (corroborating part of the findings of our previous analysis of BShift task in Section \ref{bshift_experiment}). 

\section{Control Experiments}
The main motivation behind designing a control task in probing studies is to check whether it is the representations that encode linguistic knowledge or the diagnostic classifier itself which plays a significant role in learning nuances of the task \cite{hewitt-liang-2019-designing}.
In this regard, most of our experiments throughout the paper (similarity curves, tSNE plots, or attention-norm analysis) all rely on fixed representations and do not need any classifier or training; hence, they all serve as control experiments or sanity checks. For example, in our attention-norm analysis (which requires no training and comes from a different perspective) we arrive at the same results as our attribution analysis. 


Computation of attribution scores based on trained diagnostic classifiers is the only part of our experiments which involves a training procedure. Hence, we carried out a control study inspired by \citet{talmor-etal-2020-olmpics} to check the consistency of attribution patterns.
The intuition behind this is in line with \citet{voita-titov-2020-information} who stated that if there is a strong regularity in the representations with respect to the labels, this can be revealed even with fewer training data points.

To this end, we used only 10\% of the training data to train the diagnostic classifiers and computed the attribution scores for each task. Then, we computed the correlation between attribution scores for each sentence obtained by these classifiers and those obtained from the original classifiers (trained on full training data). 
After averaging the correlations over all examples, we report the mean and maximum statistics among all layers in Table \ref{table:control_task}. 
The strong correlations imply that a similar pattern exist in the attribution scores even when fewer training instances are used.
This highlights the fact that task-specific knowledge is well encoded and regularized in the representations, nullifying the possibility of the classifier playing a major role.





\begin{table}[!t]
\centering
\resizebox{\columnwidth}{!}
{
\begin{tabular}{l  cc  cc}
\toprule
\multicolumn{1}{c}{} &
\multicolumn{2}{c}{\textbf{Pearson's $r$}} &
\multicolumn{2}{c}{\textbf{Spearman's $\rho$}}\\

\cmidrule(lr){2-3}
\cmidrule(lr){4-5}
\textbf{\textbf{Task}} & Mean & Max. & Mean & Max.\\ 
\midrule
SentLen & 0.80 & 0.97 & 0.71 & 0.94\\
ObjNum & 0.60 & 0.91 & 0.63 & 0.98\\
Tense & 0.84 & 0.93 & 0.67 & 0.91\\
BShift & 0.84 & 0.92 & 0.79 & 0.87\\
CoordInv & 0.63 & 0.90 & 0.54 & 0.85\\
\bottomrule
\end{tabular}
}
\caption{The Pearson's $r$ and Spearman's $\rho$ correlations averaged over all examples, reporting the mean and maximum values across all layers.\footnotemark
}
\label{table:control_task}
\end{table}

\footnotetext{Results are averaged over three runs.}

\section{Conclusions}
In this paper we carried out an extensive gradient-based attribution analysis to investigate the nature of BERT token representations.
To our knowledge, this is the first effort to explain probing performance results from the viewpoint of token representations.
We found that, while most of the positional information is diminished through layers, sentence-ending tokens are partially responsible for carrying this knowledge to higher layers in the model.
Furthermore, we analyzed the grammatical number and tense information throughout the model. 
Specifically, we observed that BERT tends to encode verb tense and noun number information in the \texttt{\#\#s} token and that it can clearly distinguish the two usages of the token by separating them into distinct subspaces in the higher layers.
Also, we found that abnormalities can be captured by specific token representations, e.g., in two consecutive swapped tokens or a coordinator between two swapped clauses.

Our approach in using a simple diagnostic classifier and incorporating attribution methods provides a novel way of extracting qualitative results in probing studies.
This can be seamlessly applied to various deep pre-trained models, providing a wide range of options in sentence-level tasks and from the fine-grained viewpoint of tokens.
We hope this will spur future probing studies in other evaluation scenarios. Future work might investigate how subspaces are evolved or transformed during fine-tuning and whether they are beneficial at inference time to various downstream tasks (e.g. syntactic abnormalities, grammatical number and tense subspaces in grammar-based tasks like CoLA \citealp{warstadt-etal-2019-neural}) or to check whether these behaviors are affected by different training objectives.
Furthermore, our token-level analysis can provide insights for enhancing model efficiency based on token importance, something we plan to pursue in future work.

\section*{Acknowledgments}
We would like to thank the anonymous reviewers for their helpful suggestions.

\bibliography{anthology,custom}
\bibliographystyle{acl_natbib}

\clearpage
\appendix
\counterwithin{figure}{section}
\counterwithin{table}{section}

\section{Appendices}
\label{sec:appendix}

\subsection{Probing Performance Results}
\label{sec:app_avg_rep}
Table \ref{table:avgAll} shows the results for our average sentence representation strategy (cf. Section \ref{sec:sentence-rep}) for all layers and across all tasks.
We trained the diagnostic classifiers three times and reported the expected test performance for each task \citep{dodge-etal-2019-show}.
Each task consists of 100k examples for training and 10k examples for validating the diagnostic classifiers. The test set includes 10k examples that are used for our evaluation and in-depth analysis. All dataset splits are balanced for their target classes.
The performance trends in our experiments are similar to those observed by \citet{jawahar-etal-2019-bert}.

\begin{table}[!hp]
\centering
\setlength{\tabcolsep}{12.0pt}
\renewcommand{\arraystretch}{1.1}
\resizebox{\linewidth}{!}{%
\begin{tabular}{cccccc}
\rot{\bf Layer} & \rot{\bf SentLen} & \rot{\bf BShift}                                   & \rot{\bf Tense}                                    & \rot{\bf ObjNum}                                   & \rot{\bf CoordInv}                                  \\
\bf 1 & \cellcolor[rgb]{0.992,0.906,0.145}0.91                                   & \cellcolor[rgb]{0.282,0.6,0.471}\textcolor{white}{0.57}   & \cellcolor[rgb]{0.863,0.882,0.196}0.87 & \cellcolor[rgb]{0.706,0.851,0.255}0.82 & \cellcolor[rgb]{0.267,0.576,0.482}\textcolor{white}{0.55}  \\
\bf 2 & \cellcolor[rgb]{0.925,0.894,0.173}0.89                                   & \cellcolor[rgb]{0.392,0.796,0.373}0.72 & \cellcolor[rgb]{0.894,0.886,0.184}0.88 & \cellcolor[rgb]{0.737,0.859,0.243}0.83 & \cellcolor[rgb]{0.282,0.6,0.471}\textcolor{white}{0.57}    \\
\bf 3 & \cellcolor[rgb]{0.769,0.863,0.231} \cellcolor[rgb]{0.925,0.894,0.173}0.89 & \cellcolor[rgb]{0.769,0.863,0.231}0.84                                     & \cellcolor[rgb]{0.894,0.886,0.184}0.88 & \cellcolor[rgb]{0.8,0.871,0.22} 0.84                                     & \cellcolor[rgb]{0.302,0.639,0.451}\textcolor{white}{0.60}   \\
\bf 4 & \cellcolor[rgb]{0.8,0.871,0.22} 0.85                                        & \cellcolor[rgb]{0.863,0.882,0.196}0.87 & \cellcolor[rgb]{0.925,0.894,0.173}0.89 & \cellcolor[rgb]{0.8,0.871,0.22} 0.85                                     & \cellcolor[rgb]{0.318,0.667,0.435}\textcolor{white}{0.62}  \\
\bf 5 & \cellcolor[rgb]{0.675,0.847,0.267}0.81                                   & \cellcolor[rgb]{0.894,0.886,0.184}0.88 & \cellcolor[rgb]{0.925,0.894,0.173}0.89 & \cellcolor[rgb]{0.831,0.875,0.208}0.86 & \cellcolor[rgb]{0.345,0.718,0.412}\textcolor{white}{0.66}  \\
\bf 6 & \cellcolor[rgb]{0.58,0.831,0.302}0.78                                    & \cellcolor[rgb]{0.957,0.898,0.161}0.90  & \cellcolor[rgb]{0.925,0.894,0.173}0.89 & \cellcolor[rgb]{0.831,0.875,0.208}0.86 & \cellcolor[rgb]{0.38,0.784,0.38}0.71    \\
\bf 7 & \cellcolor[rgb]{0.373,0.769,0.384}0.70                                    & \cellcolor[rgb]{0.992,0.906,0.145}0.91 & \cellcolor[rgb]{0.925,0.894,0.173}0.89 & \cellcolor[rgb]{0.831,0.875,0.208}0.86 & \cellcolor[rgb]{0.455,0.808,0.349}0.74  \\
\bf 8 & \cellcolor[rgb]{0.361,0.745,0.4}0.68                                     & \cellcolor[rgb]{0.992,0.906,0.145}0.91 & \cellcolor[rgb]{0.925,0.894,0.173}0.89 & \cellcolor[rgb]{0.8,0.871,0.22}0.85    & \cellcolor[rgb]{0.486,0.812,0.337}0.75  \\
\bf 9 & \cellcolor[rgb]{0.294,0.627,0.455}\textcolor{white}{0.59}                                   & \cellcolor[rgb]{0.992,0.906,0.145}0.91 & \cellcolor[rgb]{0.925,0.894,0.173}0.89 & \cellcolor[rgb]{0.8,0.871,0.22}0.85    & \cellcolor[rgb]{0.518,0.82,0.325}0.76   \\
\bf 10 & \cellcolor[rgb]{0.224,0.494,0.522}\textcolor{white}{0.49}                                   & \cellcolor[rgb]{0.992,0.906,0.145}0.91 & \cellcolor[rgb]{0.925,0.894,0.173}0.89 & \cellcolor[rgb]{0.769,0.863,0.231}0.84 & \cellcolor[rgb]{0.455,0.808,0.349}0.74  \\
\bf 11 & \cellcolor[rgb]{0.184,0.42,0.557}\textcolor{white}{0.43}                                    & \cellcolor[rgb]{0.957,0.898,0.161}0.90  & \cellcolor[rgb]{0.925,0.894,0.173}0.89 & \cellcolor[rgb]{0.737,0.859,0.243}0.83 & \cellcolor[rgb]{0.424,0.8,0.361}0.73    \\
\bf 12 & \cellcolor[rgb]{0.925,0.894,0.173} \cellcolor[rgb]{0.188,0.431,0.553}\textcolor{white}{0.44} & \cellcolor[rgb]{0.925,0.894,0.173} 0.89                                     & \cellcolor[rgb]{0.925,0.894,0.173} 0.89                                     & \cellcolor[rgb]{0.737,0.859,0.243}0.83 & \cellcolor[rgb]{0.392,0.796,0.373}0.72 \\
\end{tabular}
}
\caption{Layer-wise performance scores (accuracy) for the average sentence representation strategy on different probing tasks.}
\label{table:avgAll}
\end{table}

\paragraph{Mean Pooling vs. \textsc{[cls]} Pooling.}
\label{sec:appMeanVsCLS}
In order to show the reliability of our average-based pooling method for probing BERT, in Table \ref{avgvscls} we provide a comparison against the \textsc{[cls]} methodology of \citet{jawahar-etal-2019-bert}.
Specifically, we show layer-wise performance differences of the two representations, with the green color indicating improvements of our strategy.
The results clearly highlight that average representations are more suited to the task, providing improvements across many layers in most tasks.

\subsection{Full 12-layer Figures}
\label{sec:FullApp}
In this section we provide the full 12-layer version of the previous summarized layer-wise figures.

\subsection{SubjNum Mislabelling}
\label{sec:subjNumMislabel}
The SubjNum probing data suffers from numerous incorrect labels which are more obvious within samples which starts with a name that ends with an ``s'' and labelled as plural. We show five examples with this issue in Table \ref{table:incorrectLabels}.

\begin{table}[!hp]
\centering
\setlength{\tabcolsep}{12.0pt}
\renewcommand{\arraystretch}{1.1}
\resizebox{\linewidth}{!}{%
\begin{tabular}
{clllll}
\rot{\bf Layer} & \rot{\bf SentLen} & \rot{\bf BShift} & \rot{\bf Tense} & \rot{\bf ObjNum} & \rot{\bf CoordInv} \\
\bf 1 & \cellcolor[rgb]{0.937,0.976,0.957}+0.03 & \cellcolor[rgb]{0.847,0.941,0.894}+0.07 & \cellcolor[rgb]{0.882,0.953,0.918}+0.05 & \cellcolor[rgb]{0.847,0.937,0.894}+0.07 & \cellcolor[rgb]{0.965,0.988,0.976}+0.02 \\
\bf 2 & 
\cellcolor[rgb]{0.89,0.957,0.925}+0.05 & \cellcolor[rgb]{0.639,0.855,0.749}+0.17 & 
\cellcolor[rgb]{0.937,0.976,0.957}+0.03 & \cellcolor[rgb]{0.937,0.976,0.957}+0.03 & \cellcolor[rgb]{0.961,0.984,0.973}+0.02 \\
\bf 3 & 
\cellcolor[rgb]{0.737,0.894,0.82}+0.12 & \cellcolor[rgb]{0.576,0.827,0.706}+0.19 & \cellcolor[rgb]{0.965,0.988,0.976}+0.02 & \cellcolor[rgb]{0.863,0.945,0.906}+0.06 & \cellcolor[rgb]{0.957,0.984,0.969}+0.02 \\
\bf 4 & 
\cellcolor[rgb]{0.671,0.867,0.773}+0.15 & \cellcolor[rgb]{0.702,0.878,0.792}+0.14 & \cellcolor[rgb]{0.961,0.984,0.973}+0.02 & 
\cellcolor[rgb]{0.902,0.961,0.933}+0.04 &
\cellcolor[rgb]{0.902,0.961,0.933}+0.05 \\
\bf 5 & \cellcolor[rgb]{0.475,0.788,0.635}+0.24 & \cellcolor[rgb]{0.847,0.941,0.894}+0.07 & \cellcolor[rgb]{0.996,0.992,0.988}~~0.00 & \cellcolor[rgb]{0.886,0.957,0.922}+0.05 & \cellcolor[rgb]{0.933,0.973,0.953}+0.03 \\
\bf 6 & \cellcolor[rgb]{0.341,0.733,0.541}+0.3 & \cellcolor[rgb]{0.824,0.929,0.878}+0.08 & \cellcolor[rgb]{0.996,0.988,0.988} -0.01 & \cellcolor[rgb]{0.875,0.949,0.914}+0.06 & \cellcolor[rgb]{0.953,0.984,0.969}+0.02 \\
\bf 7 & \cellcolor[rgb]{0.341,0.733,0.541}+0.31 & \cellcolor[rgb]{0.831,0.933,0.882}+0.08 & 
\cellcolor[rgb]{0.996,0.992,0.992}~~0.00 & \cellcolor[rgb]{0.894,0.957,0.925}+0.05 & 
\cellcolor[rgb]{0.996,0.992,0.992}~~0.00 \\
\bf 8 & \cellcolor[rgb]{0.365,0.745,0.557}+0.29 & \cellcolor[rgb]{0.843,0.937,0.89}+0.07 & \cellcolor[rgb]{0.996,0.988,0.988}~~0.00 & \cellcolor[rgb]{0.918,0.969,0.941}+0.04 & \cellcolor[rgb]{0.996,0.98,0.98} -0.01 \\
\bf 9 & \cellcolor[rgb]{0.475,0.788,0.635}+0.24 & \cellcolor[rgb]{0.91,0.965,0.937}+0.04 & \cellcolor[rgb]{0.996,0.992,0.992}~~0.00 & \cellcolor[rgb]{0.925,0.969,0.949}+0.04 & \cellcolor[rgb]{0.992,0.961,0.957} -0.02 \\
\bf 10 & \cellcolor[rgb]{0.627,0.851,0.741}+0.17 & \cellcolor[rgb]{0.91,0.965,0.937}+0.04 & \cellcolor[rgb]{0.996,0.992,0.992}~~0.00 & \cellcolor[rgb]{0.906,0.965,0.937}+0.04 & \cellcolor[rgb]{0.984,0.929,0.925} -0.04 \\
\bf 11 & \cellcolor[rgb]{0.671,0.867,0.773}+0.15 & \cellcolor[rgb]{0.922,0.969,0.945}+0.04 & \cellcolor[rgb]{0.996,0.992,0.992}~~0.00 & \cellcolor[rgb]{0.906,0.965,0.933}+0.04 & \cellcolor[rgb]{0.984,0.929,0.925} -0.04 \\
\bf 12 & \cellcolor[rgb]{0.584,0.831,0.71}+0.19 & \cellcolor[rgb]{0.953,0.98,0.965}+0.02 & \cellcolor[rgb]{0.996,0.992,0.992}~~0.00 & \cellcolor[rgb]{0.898,0.961,0.929}+0.05 & \cellcolor[rgb]{0.988,0.949,0.945} -0.03 \\
\end{tabular}
}
\caption{Layer-wise performance scores comparison between average and \textsc{[cls]} representations across different probing tasks. Average pooling retains or improves \textsc{[cls]} performance in all layers and tasks, except for some layers in CoordInv.}
\label{avgvscls}
\end{table}

\begin{figure*}[hp]
    \includegraphics[width=\linewidth]{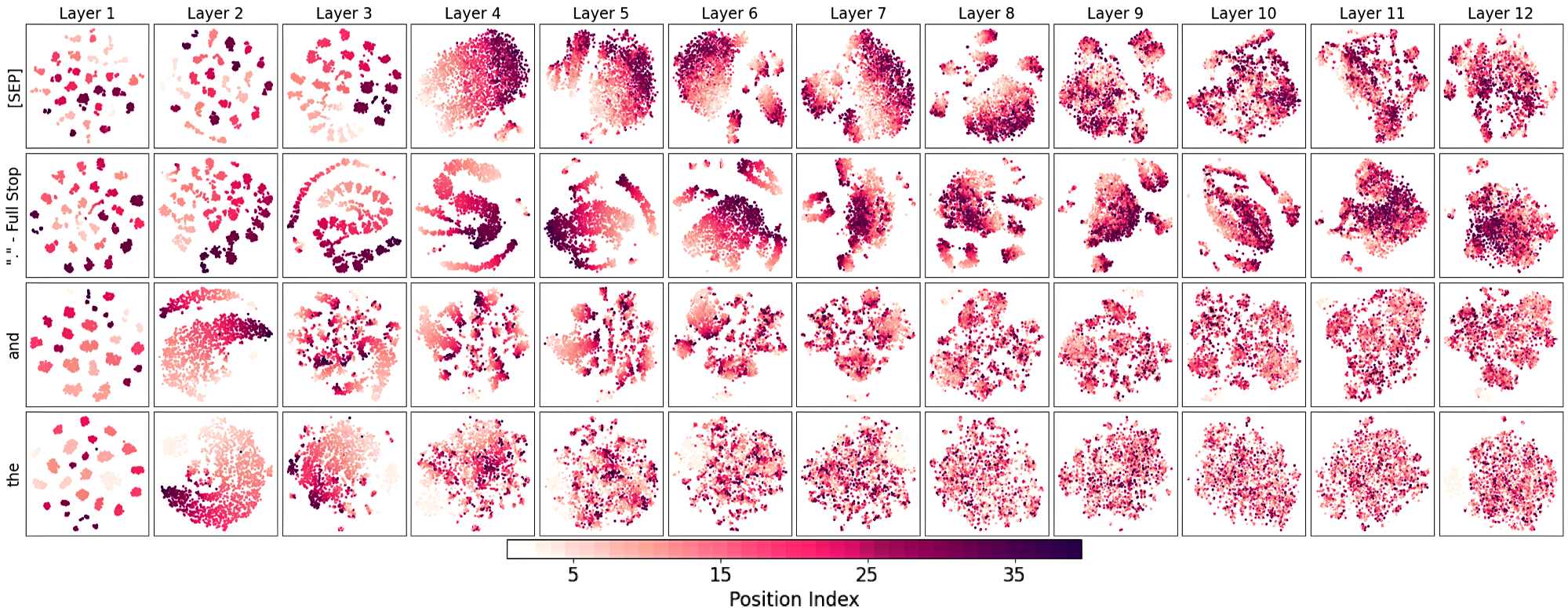}
    \caption{Full 12 layer t-SNE plots -- (Figure \ref{fig:tokenpositions})}
    \label{fig:PosEmbFullApp}
\end{figure*}

\begin{figure*}[hp]
    \includegraphics[width=\linewidth]{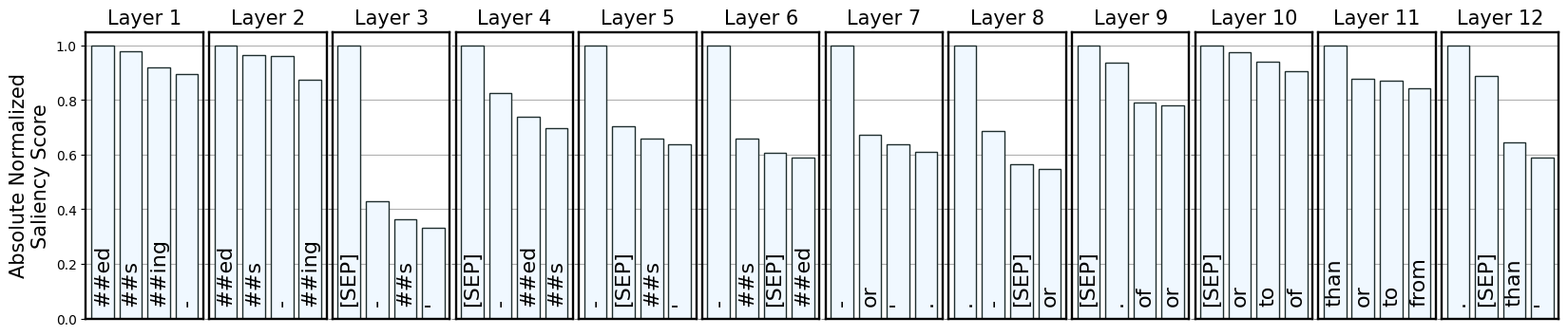}
    \caption{Full 12 layers for the top-4 most attributed high frequency tokens in the Sentence Length task - (Figure \ref{fig:senlen_bar})}
    \label{fig:senlen_bar_full}
\end{figure*}

\begin{figure*}[hp]
    \includegraphics[width=\linewidth]{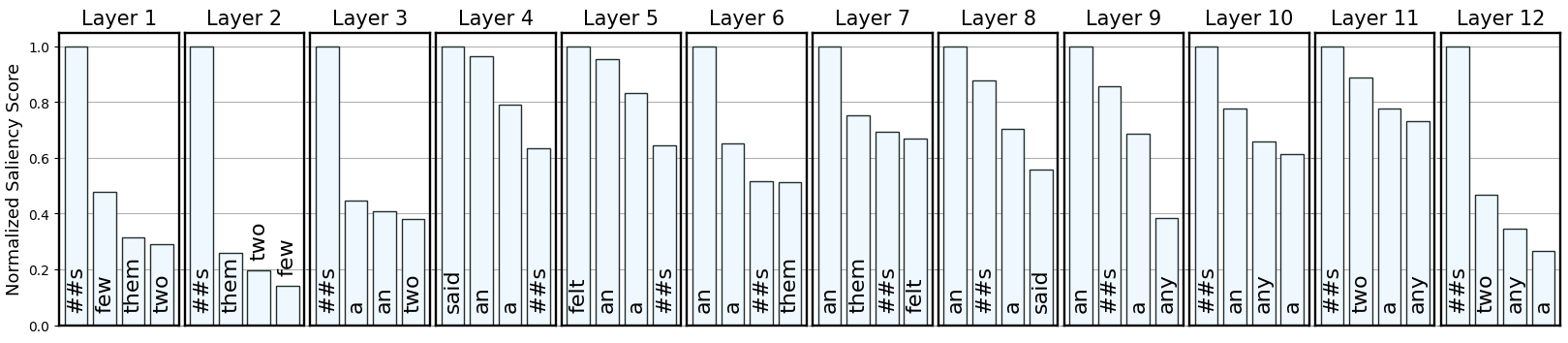}
    \caption{Full 12 layers for the top-4 most attributed high frequency tokens in the ObjNum task - (Figure \ref{fig:tenseObjNum})}
    \label{fig:objnum_bar_full}
\end{figure*}

\begin{figure*}[hp]
    \includegraphics[width=\linewidth]{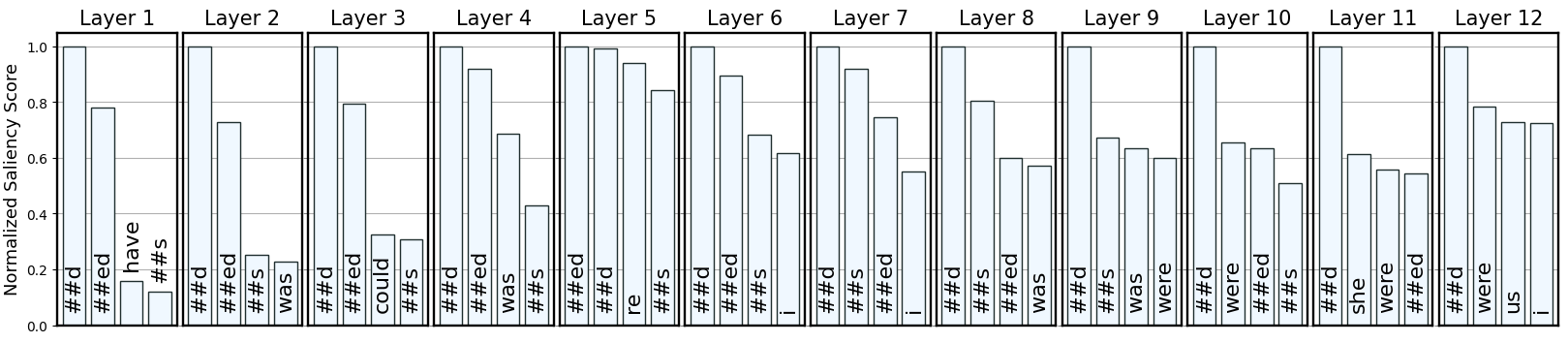}
    \caption{Full 12 layers for the top-4 most attributed high frequency tokens in the Tense task - (Figure \ref{fig:tenseObjNum})}
    \label{fig:tense_bar_full}
\end{figure*}

\begin{figure*}[hp]
    \includegraphics[width=\linewidth]{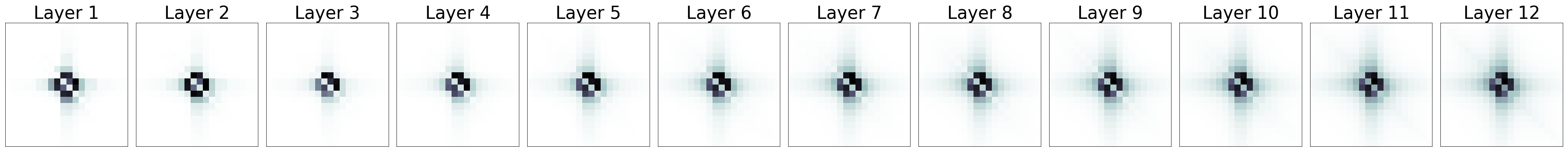}
    \caption{Full 12 layers attention-norm changes ($\Delta_{\beta^l}$) - (Figure \ref{fig:BShiftAttention})}
    \label{fig:BshiftAttention12L}
\end{figure*}

\begin{figure*}[hp]
    \includegraphics[width=\linewidth]{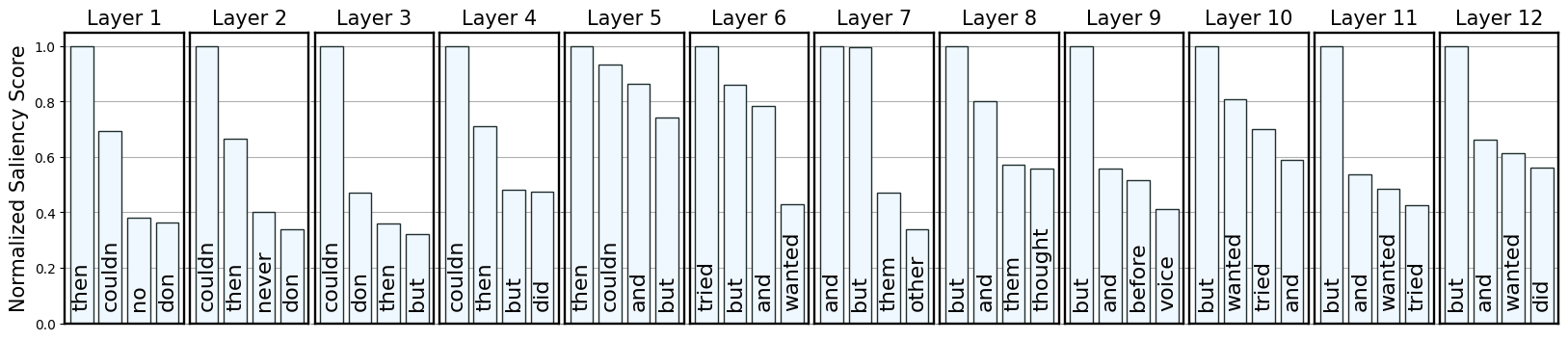}
    \caption{Full 12 layers for the top-4 most attributed high frequency tokens in the CoordInv task}
    \label{fig:coord_bar_full}
\end{figure*}

\begin{table*}
\centering
\begin{tabular}{ll}
\toprule
\textbf{Label} & \textbf{Sentence}                                                                   \\ 
\midrule
NNS   & Zeus is the child of Cronus and Rhea , and the youngest of his siblings .  \\
NNS   & Jess had never done anything this wild in her life .                       \\
NNS   & Lois had stopped in briefly to visit , but didn 't stay very long .        \\
NNS   & Tomas sank back on the seat , wonder on his face .                         \\
NNS   & Justus was an unusual man .                                                \\
\bottomrule
\end{tabular}
\caption{\label{table:incorrectLabels}
Five examples from SentEval's SubjNum data that are incorrectly labelled as plural (NNS) while the subject is clearly singular (NN). There are numerous such mislabeled instances in the test set.
}
\end{table*}

\end{document}